\pdfoutput=1

\documentclass[11pt]{article}

\usepackage[final]{acl}

\usepackage{times}
\usepackage{latexsym}

\usepackage[T1]{fontenc}

\usepackage[utf8]{inputenc}

\usepackage{microtype}

\usepackage{inconsolata}

\usepackage{graphicx}
\usepackage{caption}
\usepackage{graphicx} 
\usepackage{tikz}
\usepackage{printlen}
\usepackage{float}
\usetikzlibrary{shapes, positioning, fit}
\usepackage{amsmath}
\usepackage{algorithm}
\usepackage{algpseudocode}
\usepackage{authblk}

\title{On the Feasibility of Vision-Language Models for Time-Series Classification}

\author[ ]{%
\begin{tabular}{c c c}
    \textbf{Vinay Prithyani$^1$}\thanks{Work done outside of role at Citadel Securities} & \textbf{Mohsin Mohammed$^2$} & \textbf{Richa Gadgil$^3$} \\
    \textbf{Ricardo Buitrago$^3$} & \textbf{Vinija Jain$^4$}\thanks{Work done outside of role at Meta} & \textbf{Aman Chadha$^5$}\thanks{Work done outside of role at Amazon Web Services} \\
    \vspace{-2mm}
\end{tabular} \\
\vspace{-2mm}
\begin{tabular}{c}
    $^1$Citadel Securities \quad $^2$Telaeris Inc \quad $^3$Carnegie Mellon University\\
    $^4$Meta \quad $^5$AWS GenAI\\
\end{tabular} \\

\begin{tabular}{c}
\texttt{vinip174@gmail.com, mohsin.mohammed@telaeris.com, rgadgil@cs.cmu.edu} \\
\texttt{ricardob@andrew.cmu.edu, hi@vinija.ai, hi@aman.ai} \\
\end{tabular}\\
}

\begin{document}

\maketitle

\begin{abstract}
    We build upon time-series classification by leveraging the capabilities of Vision Language Models (VLMs). We find that VLMs produce competitive results after two or less epochs of fine-tuning. We develop a novel approach that incorporates graphical data representations as images in conjunction with numerical data. This approach is rooted in the hypothesis that graphical representations can provide additional contextual information that numerical data alone may not capture. Additionally, providing a graphical representation can circumvent issues such as limited context length faced by LLMs. To further advance this work, we implemented a scalable end-to-end pipeline for training on different scenarios, allowing us to isolate the most effective strategies for transferring learning capabilities from LLMs to Time Series Classification (TSC) tasks. Our approach works with univariate and multivariate time-series data. In addition, we conduct extensive and practical experiments to show how this approach works for time-series classification and generative labels.
    Code and datasets used in this study are available at GitHub\footnote{VLM-TSC GitHub repository: \url{https://github.com/vinayp17/VLM_TSC}}.
\end{abstract}

\section{Introduction}
Time Series Classification (TSC) is an important and challenging problem in data mining, with a variety of applications. Over the years, hundreds of time-series algorithms have been proposed, with varying success.

Recently, Large Language Models (LLMs) have gained prominence due to their abilities to capture patterns in sequential data, particularly in human language. Additionally, the modalities of vision and language have converged, resulting in the emergence of VLMs, or Vision-Language Models. One such VLM is LLAVA, which combines a vision encoder and Vicuna, an open-source LLM, for general-purpose visual and language understanding.

At the same time, LLMs have been used mainly for time-series forecasting tasks. Successful papers have framed time series forecasting as next
token prediction in text. We seek to understand how LLMs can also be adept at classifying time-series data due to sequential pattern mining. Additionally, VLMs have shown great promise on general Visual-Question Answering (VQA) tasks. By providing a graphical representation of time-series data, we aim to make LLMs more robust.

Additionally, our method requires an out-of-box VLM model to be fine-tuned for one to two epochs. This simple method provides competitive results to many deep learning frameworks, which may require longer training times. We explore how:
\begin{enumerate}
    \item Fine-tuning a VLM on time-series images and
language for two epochs or less can produce
competitive results.
    \item VLMs excel on temporal time-series data 
(such as sensor data) over spatial time-series 
data.
    \item VLMs can struggle with generalization, especially for multi-class and clustered labeltime-series.
\end{enumerate}

\section{Background}
\textbf{Time Series Classification (TSC)} TSC consist in classifying a time series into one of several possible categories. This problem arises in many areas such as medical devices, motion sensors or audio processing \cite{middlehurst2023bake}. The largest public dataset for such problem is the UCR Time Series Archive \cite{dau2018ucr} Deep Learning models have been developed for TSC, such as InceptionTime \cite{inception}, which is an ensemble of Convolutional Neural Networks (CNN) based architectures, or H-Inception \cite{ismail2022deep}, an improvement using hand-crafted filters. Other non deep learning techniques have proven to be very effective, HIVE-COTE \cite{middlehurst2021hive}, which is an ensemble of TSC models.

 \textbf{Foundation Models} Training data is limited compared to the size of today’s models \cite{fawaz2018data}, which makes overfitting a common problem. As a result, some pre-trained models have been used to tackle Time Series problems. The first layers of H-Inception has been pre-trained by using time series from different datasets and classifying the dataset to which the time series belong \cite{ismail2023finding}, and  then finetuned to each dataset individually. In the domain of Time Series Forecasting, LLMs have been shown to posses zero shot forecasting capabilities \cite{gruver2023}. However, a natural limitation of LLMs is the limited context length and the high computational costs for large sequences, which is a problem especially in Multivariate Time Series.
 
 \textbf{Vision Language Models} Several VLM have been pretrained to perform a task given a visual and  text prompt, such as LLaVA \cite{llava} and \cite{bai2024qwenvl}. These models have been trained to have general Visual Question Answering capabilities, but can be finetuned to specific datasets.

\section{Methods}

\begin{figure}[H]
    \centering
    \includegraphics[width=1.0\linewidth]{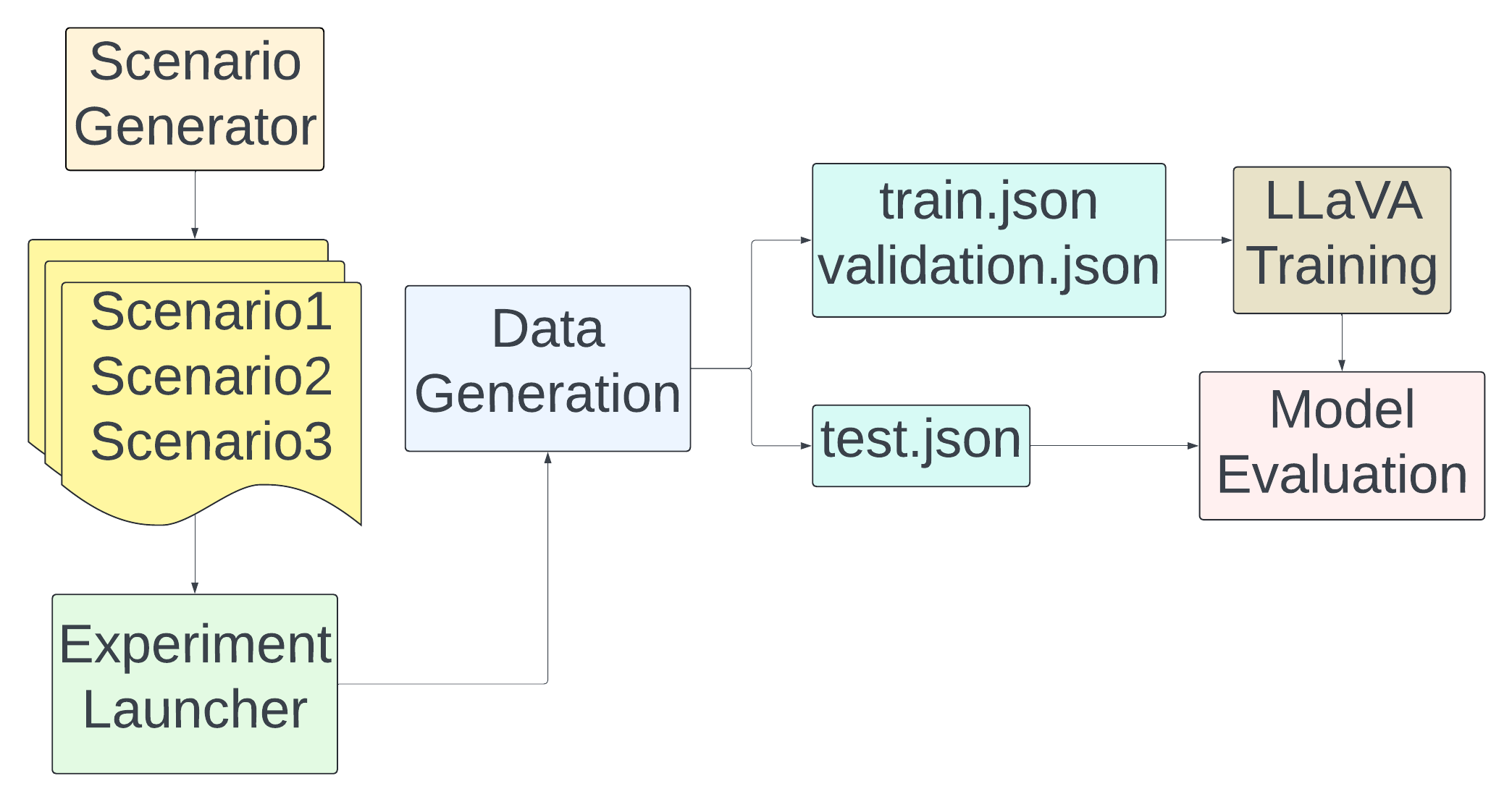}
    \caption{Pipeline Overview: Workflow for running multiple scenarios from the UCR Time Series Archive}
    \label{fig:Figure 1}
\end{figure}

\subsection{Pipeline Overview}

Our pipeline is structured into five key stages, each addressing a critical aspect of the research process: scenario generation, experiment launcher, data generation, model training, and evaluation. The following subsections describe each stage in detail:

\subsubsection{Scenario Generation}
We define a scenario to be a set of hyper-parameters, useful to test a specific hypothesis. This set includes settings for context length, down-sampling strategy, image representation type and number of epochs for training. For example in Figure 1, Scenario1 might represent a scenario with context length 2048 and an adaptive downsampling strategy. In the scenario generation phase, we generate all possible experimental scenarios to comprehensively test the hypothesis under different conditions. Our aim is to ensure a broad exploration of the experimental space by accounting for various potential influences on model performance. By systematically generating all possible scenarios, we mitigate the risk of missing critical interactions between parameters and ensure that our results are generalizable. This approach also facilitates reproducibility by documenting the full set of conditions tested.

\subsubsection{Experiment Launcher}
Primary responsibility of the experiment launcher is automating the execution of the pipeline across the set of defined datasets and different experimental scenarios. This module acts as the orchestrator, sequentially launching experiments based on the generated scenarios. Each experiment is logged, ensuring that all relevant metadata (e.g., dataset used, model configuration, hyperparameters) are captured. This automation significantly reduces human intervention, allowing for high-throughput experimentation and consistent application of the experimental protocol across different trials. It ensures that each experiment is conducted in a controlled, systematic manner, minimizing variability that could otherwise skew results.

\subsubsection{Data Generation}
During this phase, the pipeline generates the appropriate data splits—train, validation, and test—based on the chosen dataset and the scenario parameters. We employ a 80/10/10 split of the dataset, allocating 80\% of the data for training, 10\% of the data for validation and the remainder 10\% for testing. This balanced split ensures that there is sufficient data for model optimization, while reserving an adequate portion for evaluation and generalization assessment on unseen data. 
The data generated can be one of two modes: \textbf{BASELINE} or \textbf{WITH\_STATS}. BASELINE uses a comma separated list of the original time-series signal downsampled to fit within the context window as the primary source of textual features. For the mode WITH\_STATS, the pipeline augments BASELINE data by computing and appending statistical features (e.g mean, variance, skewness, kurtosis ) derived from the original time series signal. These statistical features provide additional context and information about the signal's characteristics, leading to a richer input representation for the model.

\begin{algorithm}
\caption{Data Splitting and Feature Generation}
\begin{algorithmic}
\Require Dataset $D$, Scenario parameters $S$, Mode $M \in \{\text{BASELINE}, \text{WITH\_STATS}\}$
\Ensure Training, validation, and test splits with corresponding feature representations
\State Split dataset $D$ into:
\State \hspace{\algorithmicindent} training set $D_{\text{train}}$ (80\%), 
\State \hspace{\algorithmicindent} validation set $D_{\text{val}}$ (10\%), and
\State \hspace{\algorithmicindent} test set $D_{\text{test}}$ (10\%) based on scenario parameters $S$.
\For{each subset $D_{\text{sub}} \in \{D_{\text{train}}, D_{\text{val}}, D_{\text{test}}\}$}
    \For{each time-series sample in $D_{\text{sub}}$}
        \State Generate a comma-separated list of signal values, downsampled to fit the context window.
        \If{$M = \text{WITH\_STATS}$}
            \State Compute additional statistical features (e.g., mean, standard deviation, min, max).
            \State Append these features to the string representation.
        \EndIf
    \EndFor
\EndFor
\State \Return Data splits $D_{\text{train}}$, $D_{\text{val}}$, $D_{\text{test}}$ with generated features
\end{algorithmic}
\end{algorithm}

\subsubsection{Training}
\begin{figure}[H]
    \centering
    \includegraphics[width=1.0\linewidth]{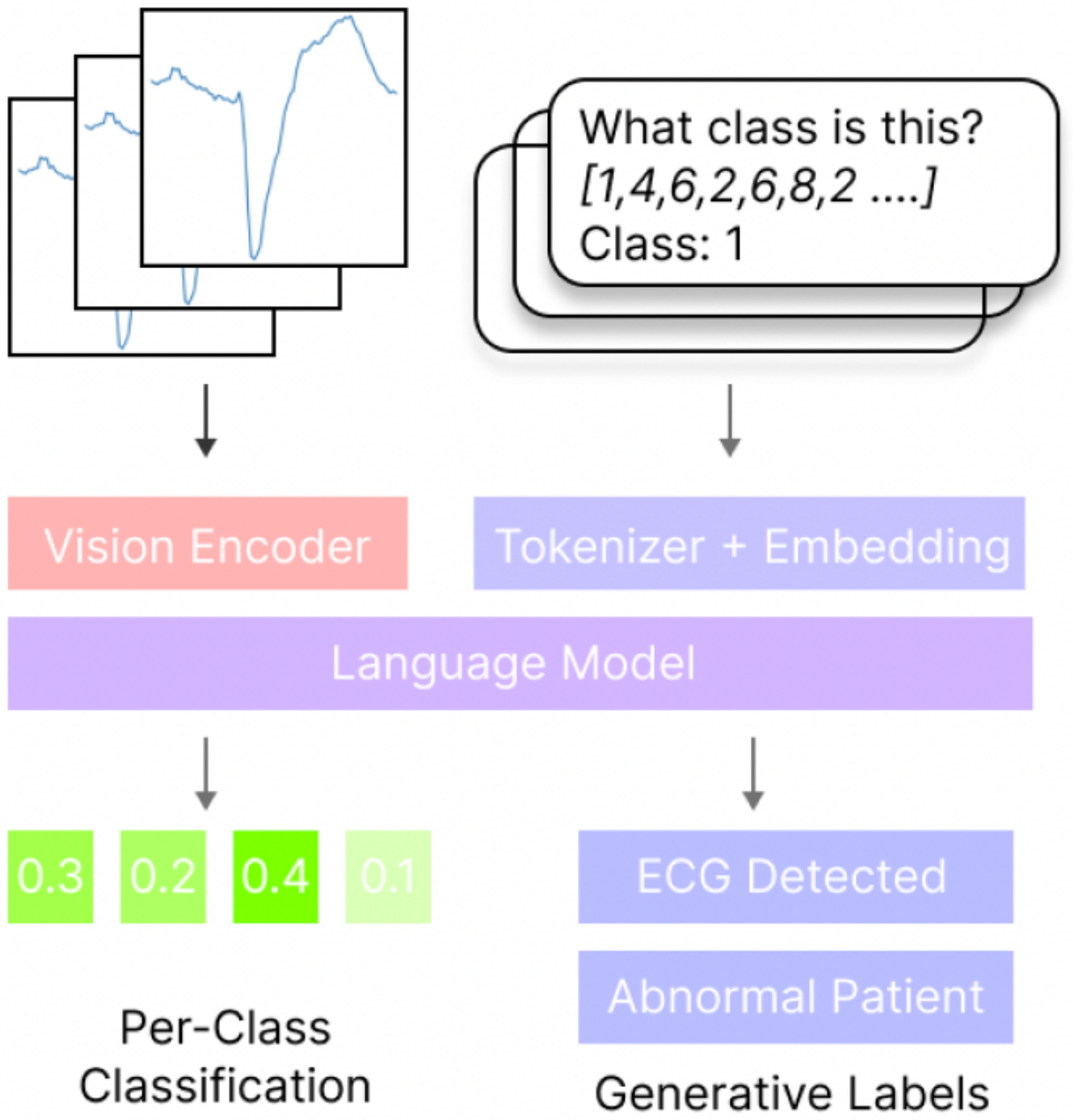}
    \caption{System Architecture: The top left depicts Time Series images fed into the Vision Encoder, combined with a tokenized Text Prompt, and both processed through a shared language model (Vicuna) to perform Time Series Classification}
    \label{fig:Figure 2}
\end{figure}

LLaVA's groundbreaking architecture (Figure \ref{fig:Figure 2}) combines the capabilities of pre-trained language models like Vicuna or LLaMA with visual encoders like CLIP. This fusion is achieved by converting the visual features extracted from images into a format compatible with the language model’s embeddings. To facilitate this alignment, a trainable projection matrix is utilized, generating a sequence of visual token embeddings that can be seamlessly integrated into the language model.
This stage performs fine-tuning of the LLaVA model (Language and Vision Assistant) using the generated training data. The model is trained based on the scenario-specific parameters, which may include adjustments to number of epochs, context length. 

\subsubsection{Evaluation}
The final stage evaluates the fine-tuned model on unseen test data, specifically focusing on one-shot accuracy. One-shot accuracy measures the model's ability to generalize from minimal training examples, making it an essential metric in few-shot learning tasks. The results from the evaluation stage are recorded and analyzed, providing insights into the model's performance under different conditions. The evaluation process offers a rigorous assessment of the model's generalization capabilities. By focusing on one-shot accuracy, we aim to measure the model's effectiveness in real-world applications where access to large amounts of labeled data is often limited. The results from this stage directly inform conclusions about the model's suitability for deployment in similar environments.

\subsection{Graphical and Text-Based Representations of Time-Series Data}

\begin{figure}[H]
    \centering
    \begin{tikzpicture}
        \node[draw, rounded corners, align=left, text width=7.5cm, minimum height=3cm, font=\large] (box) {
            \textbf{Which class is the following signal from?} \\
            $2, 3, 4, 5,10, 3 \dots$ \\
            \textbf{Class:}
        };
    \end{tikzpicture}
    \caption{Univariate Baseline Prompt Template Example: Prompt consists of a question followed by a comma-separated list of the signal}
    \label{fig:Figure 3}
\end{figure}

\begin{figure}[H]
    \centering
    \begin{tikzpicture}
        \node[draw, rounded corners, align=left, text width=7.5cm, minimum height=3cm, font=\large] (box) {
            \textbf{Which class is the following signal from?} \\
            $2, 3, 4, 5,10,3, Mean:4.5 Variance:6.91 \dots$ \\
            \textbf{Class:}
        };
    \end{tikzpicture}
    \caption{Univariate With Stats Prompt Template Example: Prompt consists of a question followed by a comma-separated list of the signal and ending with basic statistics calculated on the signal}
    \label{fig: Figure 4}
\end{figure}

\begin{figure}[H]
    \centering
    \begin{tikzpicture}
        \node[draw, rounded corners, align=left, text width=7.5cm, minimum height=3cm, font=\large] (box) {
            \textbf{Which class is the following signal from?} \\
            {Dimension One}: $2, 3, 4, 5, 10, 3 \dots$ \\
            {Dimension Two}: $4, 2, 4, 3, 1, 3  \dots$ \\
            \dots \\
            {Dimension N}: $2, 6, 4, 9, 90, 3 \dots$ \\

            \textbf{Class:}
        };
    \end{tikzpicture}
    \caption{Multivariate Baseline Prompt Template Example: Prompt consists of a question followed by a comma-separated list of the signal across each dimension}
    \label{fig: Figure 5}
\end{figure}

For LLMs, we construct the prompt in a customizable fashion based on the following:
\begin{enumerate}
    \item Univariate or Multivariate signal
    \item BaseLine or With Stats
\end{enumerate}

As can be seen in Figures~\ref{fig:Figure 2}-\ref{fig: Figure 4}, the \textit{Baseline} prompt simply consists of the time series signal. The \textit{With Stats} prompt further enhances the prompt with statistics calculated from the time series signal. For multi-dimensional time series data, we enumerate the signal and statistics across each dimension.

The integration of graphical representations in conjunction with text-based data is a cornerstone of our approach to enhancing the        capabilities of Vision-Language Models (VLMs) in interpreting time-series data. For graph data, our primary focus is on generating clean, uncluttered line plots that effectively convey the time-series data without extraneous elements. This decision is based on the understanding that additional features in a graph, such as titles, axis labels,   and legends, while informative in a human-readable context, can introduce unnecessary complexity and noise when processed by a VLM. These elements could potentially distract the model from the core data trend or pattern that the line plot is intended to represent.

\subsection{Model Context Length Variation}
In our exploration of the capabilities of LLMs and VLMs, a key aspect we investigate is the impact of varying the context length on model performance. Specifically, we categorize datasets into two groups based on their alignment with our total context length, which is set at 2048 tokens. This threshold is significant as it represents a common    constraint in many language models, where the capacity to process and interpret information is bounded by a maximum token limit. For datasets that exceed this 2048-token limit, we employ different down-sampling strategies. Down-sampling involves reducing the size of the dataset to fit within the prescribed token limit, a process that inherently leads to a loss or corruption of some data. Aside from context windows, this approach simulates scenarios where the available data is in- complete or partially corrupted, a common challenge in real-world applications.

We do this by applying different downsampling strategies. The prupose of this filtering step is to smooth out the signal. We test whether the incorporation of a visual component can compensate for the loss of information due to downsampling. By introducing visual data, we aim to ascertain if the VLM can leverage this additional modality to maintain or even enhance it's performance despite the reduced textual data. 

\subsection{Downsampling Strategies}

\subsubsection{Uniform Downsampling}
Uniform downsampling (Figure \ref{fig:Figure 7}) is a technique used in signal processing to reduce the sampling rate of a time series by selecting data points at regular intervals. Unlike adaptive downsampling, which adjusts the sampling rate based on the variability of the signal, uniform downsampling treats all sections of the signal equally, discarding intermediate points to achieve a lower resolution. This approach is particularly useful for reducing the computational burden or storage requirements when working with large datasets, while retaining the overall structure and key features of the original signal. However, it may lead to information loss in regions where the signal exhibits rapid changes, as the uniform selection does not account for variations in the data's complexity.

\begin{figure}[H]
    \centering
    \includegraphics[width=1.0\linewidth]{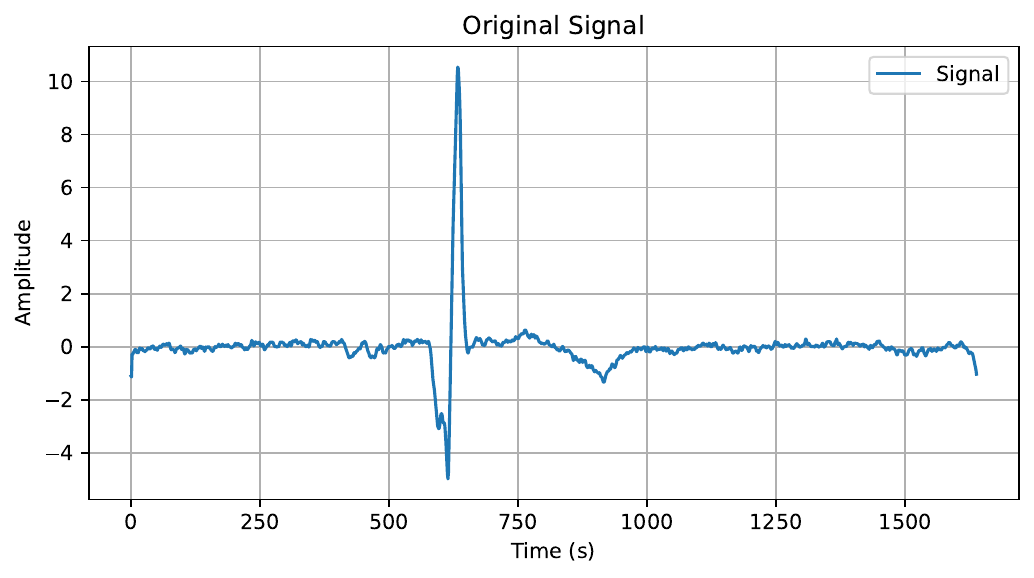}
    \caption{Original ECG image from CinCECGTorso dataset, representing multi-lead ECG recordings useful for characterizing different heart conditions.No downsampling is done on this signal}
    \label{fig:Figure 6}
\end{figure}

\begin{figure}[H]
    \centering
    \includegraphics[width=1.0\linewidth]{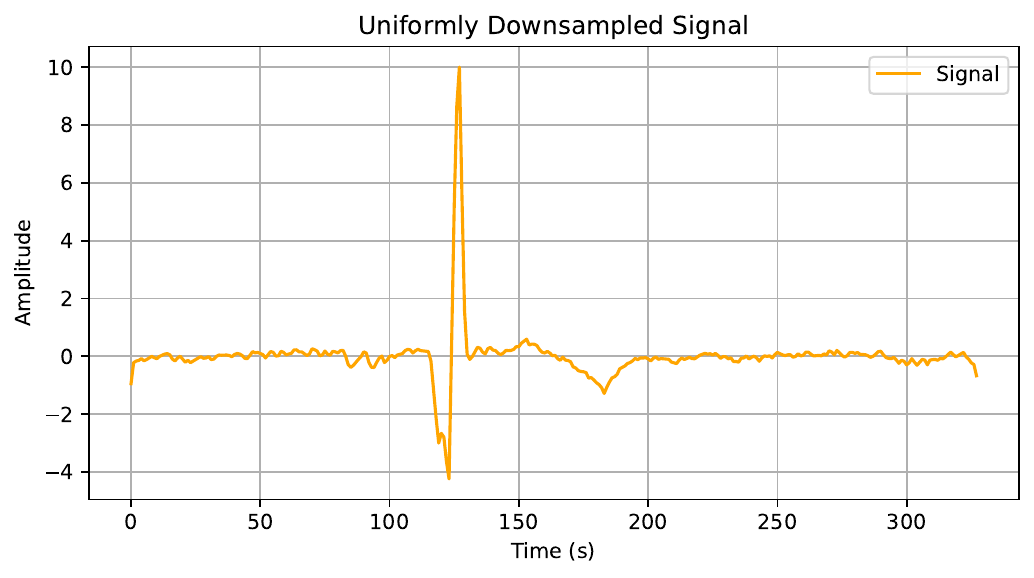}
    \caption{Original ECG signal from CinCECGTorso dataset, is passed through a uniform filter, equally reducing the data-points while still maintaining the original shape to fit within context-length limitations}
    \label{fig:Figure 7}
\end{figure}

\subsubsection{Adaptive Downsampling}
\begin{figure}[H]
    \centering
    \includegraphics[width=1.0\linewidth]{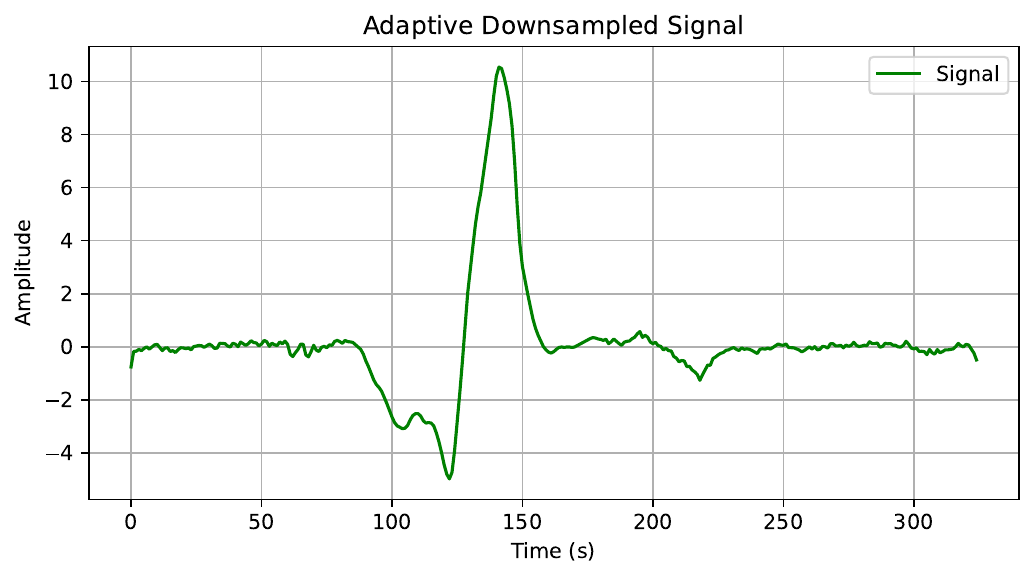}
    \caption{Original ECG signal from CinCECGTorso dataset, is adaptively downsampled, reducing more data in parts of the signal with low variability and preserving more data in parts of the signal with high variability}
    \label{fig:Figure 8}
\end{figure}

Adaptive downsampling (Figure \ref{fig:Figure 8}) is a technique used to reduce the sampling rate of a time series signal in a way that preserves important information, particularly in regions of the signal that exhibit high variability. Unlike uniform downsampling, which samples data points at regular intervals, adaptive downsampling dynamically adjusts the rate based on the signal's characteristics. Areas of high complexity or rapid change are sampled more frequently, while regions of lower variability are sampled less densely. This approach helps to maintain the critical features of the signal, minimizing information loss, while reducing the amount of data that needs to be processed or stored. Adaptive downsampling is especially useful in scenarios where the data exhibits non-uniform behavior, and efficient compression is needed without compromising signal integrity.

\begin{algorithm}[H]
\caption{Adaptive Downsampling Overview}
\begin{algorithmic}
\Require Input signal $X = \{x_1, x_2, ..., x_n\}$, threshold $\tau$, window size $w$, downsampling factor $f$ and lambda ${adaptive\_downsampler\_impl}({variability}, \tau)$
\Ensure Downsampled signal $Y$
\State Initialize empty list $Y$
\State $\texttt{target\_len} = \frac{\texttt{Len}(X)}{f}$
\State $\texttt{number\_windows} = \frac{\texttt{Len}(X)}{w}$
\State Initialize empty list $variability$
\For{$i= 0$ to $num\_windows$}
    \State $window = X[i*w:(i+1)*w]$
    \State ${variability.append}({stddev}(window))$
\EndFor
\State $Y = adaptive\_downsample\_impl(variability, \tau)$
\State \textbf{return} Downsampled signal $Y$
\end{algorithmic}
\end{algorithm}

\section{Evaluation Metrics}
We focus on measuring model performance by following the paradigm of multiple-choice benchmarks for LLMs. We follow the method set by the original MMLU implementation by only extracting the predictions or predicted probabilities of the model that correspond to the answers. We do this in a one-shot setting in order to be mindful of  the context length of the model.

\section{Experiments}
\begin{figure}[h]
    \centering
    \includegraphics[width=1.0\linewidth]{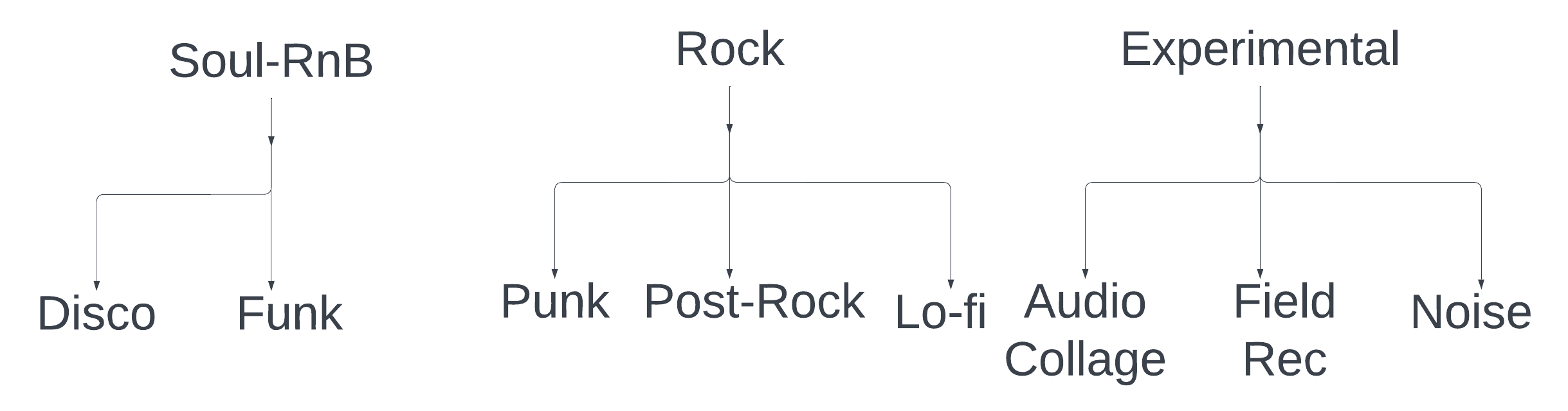}
    \caption{Clustering of FMA Genre , for eg Punk, Post-Rock, Lo-fi all belong to the parent genre Rock}
    \label{fig:Figure 9}
\end{figure}

For the single-class setting, we primarily use the UCR time-series benchmark datasets that are a widely utilized benchmark in time-series analysis and deep learning research. It provides a standardized framework for evaluating the performance of time-series classification algorithms across a diverse range of domains. For the multi-class setting, we consider two datasets. The first dataset is The Free Music Archive (FMA). The dataset (Figure \ref{fig:Figure 9}) includes 106,574 tracks categorized into a hierarchical structure of 161 genres. We leverage the hierarchical nature   of the genres to construct a multi-label dataset. For this setting, we cluster the labels within each parent genre and have the model predict the parent genre, as a way of doing future extreme label classification. Additionally, it contains temporal features for each track collected by EchoNet, that we use as our uni-variate time-series.

Finally, we consider a multi-variate, multi-class dataset in the form of sensor values from a hydraulic test rig.    The system measures the condition of four hydraulic components (cooler, valve, pump and accumulator). As our label,    we pass in the status of each component, and attempt to prompt the model by appending the component name to the input  during evaluation. 

\textbf{Models} We use LLAVA for our pre-trained VLM model, with CLIP Vit Large Patch as the Vision Encoder    and Vicuna-7B as the Large Language Model. 

\textbf{Training} We train each LLAVA model on one A100 GPU for two epochs, using lora\_r as 128 and lora\_alpha as 256.

\section{Results}

\subsection{A/B Testing Line Plots vs. Scatter Plots}
In our exploration, we also aim to study the impact, different graphical representations have on 
Time Series Classification tasks. For this purpose, we conducted an A/B test comparing two types of graphical representations: line plots and scatter plots. The goal is to identify which representation better aided the model in capturing relevant temporal patterns in the data resulting in better one-shot accuracy. 

\subsubsection{Experiment Setup}

We selected the following datasets from the UCR Time Series Archive.
\begin{itemize}
    \item \textbf{ItalyPowerDemand}: Time series depicting daily power demand 
    \item \textbf{PenDigits}: Time series depicting x,y coordinates of pen traced across a digital screen
\end{itemize}

\textbf{Graphical Representation Types}:
\begin{itemize}
    \item \textbf{Line Plot}: In this representation, all data points are connected to form a continuous line, highlighting the trend, direction, and changes over time.
    \item \textbf{Scatter Plot}: In contrast, scatter plots presented individual data points without connecting lines, treating each point as an independent entity.
\end{itemize}

\subsubsection{Hypothesis}

We hypothesized that line plots would outperform scatter plots in classification tasks due to their inherent ability to better capture the structure of time-series data due to the following reasons:
\begin{enumerate}
    \item \textbf{Temporal Structure}: Line plot continuity should make it easier for the model to detect trends and relationships between consecutive data points. 
    \item \textbf{Noise Reduction}: Isolated points in scatter plots can appear more randomly distributed, making it difficult for the model to grasp underlying patterns.
    \item \textbf{Trend Detection}: Line plots would offer a clearer visual representation of trends, periodicity, and other temporal features, which are critical for accurately classifying time-series data.
    \item \textbf{Visual Clarity for Long Sequences}:
When dealing with longer time-series sequences, scatter plots can become increasingly difficult to interpret, as the number of points grows and overlaps. Line plots, however, maintain clarity even with longer sequences, as the continuous lines provide a clear visual path through the data, aiding the model in recognizing temporal relationships.

\end{enumerate}

\subsubsection{Results}

Our findings confirmed the hypothesis. In our experiments, as shown in Table~\ref{tab:A_B_test}, line plots led to significantly higher classification accuracy compared to scatter plots, when keeping all other parameters constant. For example, in the \textbf{PenDigits} dataset, the model's accuracy with line plots was \textbf{85.08\%}, while with scatter plots, it dropped to \textbf{80.64\%}. Similar results were observed for \textbf{ItalyPowerDemand}

\begin{table}[h]
    \centering
        \begin{tabular}{ccccc}
            \hline
            \textbf{Dataset} & \textbf{Plot Type} & \textbf{Accuracy} \\ \hline
            PenDigits & Line  & 85.08\% \\ 
            PenDigits & Scatter & 80.64 \% \\ 
            ItalyPowerDemand & Line  & 95 \% \\ 
            ItalyPowerDemand & Scatter & 89.31\% \\ \hline            
        \end{tabular}
    \caption{Results of the A/B test experiments with different plot types, using uniform down-sampling and context length of 4096}
    \label{tab:A_B_test}
\end{table}

Our A/B testing showed that line plots are a more effective graphical representation for time-series classification using Vision-Language Models like LLaVA. The continuous nature of line plots, their ability to reduce noise, and their clear depiction of trends and temporal relationships contributed to their superior performance over scatter plots. These findings suggest that preserving the temporal continuity of time-series data in graphical form is critical for maximizing the performance of VLMs in TSC tasks.All subsequent experiments are done using a Line plot.

Future work may explore additional types of visualizations, such as candlestick charts or heatmaps, to further enhance the model's ability to capture complex temporal patterns. Additionally, incorporating hybrid approaches that combine the strengths of line plots with other visual representations could be investigated to further boost classification accuracy.

\subsection{Pipeline Results}
On single-label classification, as shown in Table~\ref{tab:uni_2048}, LLAVA performs quite well on most of the present data despite down-sampling. However, it should be noted that the model does best on signal data that is inherently temporal. Temporal signals, such as the ones in our power demand and ECG datasets, have inherent time-based patterns. The two datasets that display poor performance, PhalangesOutlines and Hand- Outlines (Table~\ref{tab:multi}) contain extracted hand and bone outlines, as compared  to temporal data. LLAVA may be less adept at handling static, spatial data or the specific feature representations derived from out-lines.

\begin{table}[h]
    \centering
    \resizebox{\columnwidth}{!}{
        \begin{tabular}{ccccc}
            \hline
            \textbf{Dataset} & \textbf{Context Length} & \textbf{Accuracy} \\ \hline
            CinCECGTorso & 2048 & 76.4\% \\ 
            ItalyPowerDemand & 2048 & 90.0 \% \\
            FreezerSmallTrain & 2048  & 97.39 \% \\ 
            PhalangesOutlinesCorrect & 2048 & 68.79\% \\ 
            TwoLeadECG & 2048 & 99.1\% \\  \hline
        \end{tabular}
    }
    \caption{Results of the experiments with different Univariate datasets,using a Baseline prompt type and uniform downsampling.}
    \label{tab:uni_2048}
\end{table}

\begin{table}[h]
    \centering
        \begin{tabular}{ccccc}
            \hline
            \textbf{Dataset} & \textbf{Context Length} & \textbf{Accuracy} \\ \hline
            PenDigits & 2048 & 72.89\% \\ 
            HandOutlines & 2048 & 66.7\% \\ \hline
        \end{tabular}
    \caption{Results of the experiments with different Multivariate datasets, using a Baseline prompt type, and uniform down-sampling}
    \label{tab:multi}
\end{table}

Experiments with longer context lengths ( Table ~\ref{tab:long_ctx_uni} ) improve VLM task performance across the board for high-dimensional datasets. FreezerSmallTrain is inherently lower dimensional and therefore sees only marginal performance gains. For high-dimensional datasets like CinCECGTorso and PenDigits, this leads to significant accuracy gains (76.4\% → 98.3\%, 72.89\% → 85.08\%). In PenDigits, characters or numbers represented by digitized pen strokes may span multiple timesteps. A longer context ensures the model sees the full sequence of strokes, capturing the holistic shape and flow needed for accurate classification. In CinCECGTorso, the ECG signal's temporal dependencies across longer time spans are critical for accurate classification. A limited context might miss important signal patterns that manifest over time, whereas longer contexts provide a comprehensive view of the temporal features.

\begin{table}[h]
    \centering
    \resizebox{\columnwidth}{!}{
        \begin{tabular}{ccccc}
            \hline
            \textbf{Dataset} & \textbf{Context Length} & \textbf{Accuracy} \\ \hline
            CinCECGTorso  & 4096 & 98.3\% \\ 
            PenDigits  & 4096 & 85.08\% \\ 
            FreezerSmallTrain & 4096 &  98.78 \% \\ \hline
        \end{tabular}
        }
    \caption{Results of the experiments with a Baseline prompt type, uniform downsampling and longer context length}
    \label{tab:long_ctx_uni}
\end{table}

 The model struggles the most in the multi-class setting (Table~\ref{tab:multi} ). For the FMA dataset, this indicates the model may be brittle  to clustering. LLAVA may struggle when data points within the same class are not well-clustered or when there is overlap between classes in the feature space. For the hydralics dataset, or multi-class setting without clustering,  our prompting strategy likely led to the poor performance.

 \begin{table}[h]
    \centering
    \resizebox{\columnwidth}{!}{
        \begin{tabular}{ccccc}
            \hline
            \textbf{Dataset} & \textbf{Prompt Type} & \textbf{Downsampling Strategy} & \textbf{Context Length} & \textbf{Accuracy} \\ \hline
            FMA & Baseline & Uniform & 2048 & 21.0\% \\ 
            Hydraulics & Baseline & Uniform & 2048 & 9.1\% \\ \hline
        \end{tabular}
    }
    \caption{Multiclass TimeSeries : Results of the experiments with different datasets, prompt types, downsampling strategies, and context lengths.}
    \label{tab:multi_class}
\end{table}

 It should be noted that for the multi-class setting, due to increased variance within each cluster, fine- tuning the model for less than two epochs also may have led to sub-par results.

 Our experiments with the adaptive downsampling technique (Table~\ref{tab:adaptive}) demonstrated that it was a superior method for constructing prompts, particularly for signals with fluctuating levels of variability. By preserving more detailed information in regions of higher variability, this technique provided a better signal representation compared to uniform downsampling, which often resulted in performance degradation. For CinCECGTorso we can see performance improvement from (76.4\% → 82.4\%)

 \begin{table}[h]
    \centering
    \resizebox{\columnwidth}{!}{
        \begin{tabular}{ccccc}
            \hline
            \textbf{Dataset} & \textbf{Context Length} & \textbf{Accuracy} \\ \hline
            CinCECGTorso  & 2048 & 82.4\% \\ \hline
        \end{tabular}
        }
    \caption{Results of the experiments with a Baseline prompt type, and adaptive downsampling}
    \label{tab:adaptive}
\end{table}

The comparison between the BASELINE and WITH\_STATS (Table ~\ref{tab:with_stats} data representations revealed that augmenting signals with auxiliary statistics improved model performance, particularly for more complex datasets. However, adding too many statistical features led to diminishing returns, suggesting that there is a trade-off between capturing relevant features and overwhelming the model with excessive data.

\begin{table}[h]
    \centering
    \resizebox{\columnwidth}{!}{
        \begin{tabular}{ccccc}
            \hline
            \textbf{Dataset} & \textbf{Prompt Type} & \textbf{Accuracy} \\ \hline
            CinCECGTorso  & WITH\_STATS & 79.6\% \\ \hline
        \end{tabular}
        }
    \caption{Results of the experiments with uniform down-sampling and augmenting text with summary statistics }
    \label{tab:with_stats}
\end{table}

\section{Conclusion}
In this work, we explored the potential of Vision-Language Models (VLMs) for Time Series Classification (TSC) tasks by leveraging their ability to process both graphical and textual representations of time-series data using LLaVA.  Our experiments demonstrated that VLMs can produce competitive results even with minimal fine-tuning, especially when incorporating visual representations that provide additional contextual information beyond what numerical data alone can capture.

We also introduced a scalable, end-to-end pipeline that allows for experimentation across multiple training scenarios, providing insights into the effectiveness of different configurations, such as varying context lengths, downsampling strategies, and data representations. Notably, the introduction of adaptive downsampling, which dynamically adjusts the level of downsampling based on signal variability, showed promising results by retaining more relevant information in high-variability regions.

Our findings underscore three key observations:

\begin{enumerate}
    \item Higher context lengths significantly improve performance for high-dimensional signals: Longer context lengths enable the model to capture more extensive temporal dependencies and richer feature representations, which are crucial for high-dimensional data. For example, in tasks like ECG classification, where signal patterns span longer intervals, the ability to observe a broader context led to drastic improvements in accuracy. 
    \item Choosing the right downsampling strategy is essential: The choice of downsampling strategy greatly influences model performance, as it determines how much information is preserved during preprocessing. Adaptive downsampling, which adjusts based on signal variability, was particularly effective in retaining critical information from high-variability regions while reducing noise in lower-variability segments. 
    \item Custom features computed on the original signal can boost performance: Adding engineered features derived from the original signal, such as statistical summaries or domain-specific metrics, can provide complementary information that enhances the model’s ability to discriminate between classes.
\end{enumerate}

Future work may focus on improving model generalization, particularly in multi-class settings, and further refining the adaptive methods to better handle the complexity of real-world time-series data.


\clearpage

\bibliography{custom}

\begin{thebibliography}{10}
\providecommand{\natexlab}[1]{#1}

\bibitem[{Bai et~al.(2024)Bai, Bai, Yang, Wang, Tan, Wang, Lin, Zhou, and Zhou}]{bai2024qwenvl}
Jinze Bai, Shuai Bai, Shusheng Yang, Shijie Wang, Sinan Tan, Peng Wang, Junyang Lin, Chang Zhou, and Jingren Zhou. 2024.
\newblock \href {https://openreview.net/forum?id=qrGjFJVl3m} {Qwen-{VL}: A versatile vision-language model for understanding, localization, text reading, and beyond}.

\bibitem[{Dau et~al.(2018)Dau, Bagnall, Kamgar, Yeh, Zhu, Gharghabi, Ratanamahatana, and Keogh}]{dau2018ucr}
Hoang~Anh Dau, Anthony Bagnall, Kaveh Kamgar, Chin-Chia~Michael Yeh, Yan Zhu, Shaghayegh Gharghabi, Chotirat~Ann Ratanamahatana, and Eamonn Keogh. 2018.
\newblock The ucr time series archive.
\newblock \emph{arXiv preprint arXiv:1810.07758}.

\bibitem[{Fawaz et~al.(2018)Fawaz, Forestier, Weber, Idoumghar, and Muller}]{fawaz2018data}
Hassan~Ismail Fawaz, Germain Forestier, Jonathan Weber, Lhassane Idoumghar, and Pierre-Alain Muller. 2018.
\newblock Data augmentation using synthetic data for time series classification with deep residual networks.
\newblock \emph{Proceedings of the ECML PKDD 2018 Workshop on Advanced Analytics and Learning on Temporal Data}.

\bibitem[{Fawaz et~al.(2019)Fawaz, Lucas, Forestier, Pelletier, Schmidt, Weber, Webb, Idoumghar, Muller, and Petitjean}]{inception}
Hassan~Ismail Fawaz, Benjamin Lucas, Germain Forestier, Charlotte Pelletier, Daniel~F. Schmidt, Jonathan Weber, Geoffrey~I. Webb, Lhassane Idoumghar, Pierre-Alain Muller, and François Petitjean. 2019.
\newblock Inception time.
\newblock Available at: \url{https://arxiv.org/abs/1909.04939}.

\bibitem[{Gruver et~al.(2024)Gruver, Finzi, Qiu, and Wilson}]{gruver2023}
Nate Gruver, Marc Finzi, Shikai Qiu, and Andrew~Gordon Wilson. 2024.
\newblock Large language models are zero-shot time series forecasters.
\newblock Availabl at: \url{https://arxiv.org/pdf/2310.07820}.

\bibitem[{Ismail-Fawaz et~al.(2023)Ismail-Fawaz, Devanne, Berretti, Weber, and Forestier}]{ismail2023finding}
Ali Ismail-Fawaz, Maxime Devanne, Stefano Berretti, Jonathan Weber, and Germain Forestier. 2023.
\newblock Finding foundation models for time series classification with a pretext task.
\newblock Available at: \url{https://arxiv.org/abs/2307.13172}.

\bibitem[{Ismail-Fawaz et~al.(2022)Ismail-Fawaz, Devanne, Weber, and Forestier}]{ismail2022deep}
Ali Ismail-Fawaz, Maxime Devanne, Jonathan Weber, and Germain Forestier. 2022.
\newblock Deep learning for time series classification using new hand-crafted convolution filters.
\newblock In \emph{2022 IEEE International Conference on Big Data (Big Data)}, pages 972--981. IEEE.

\bibitem[{Liu et~al.(2023)Liu, Li, Wu, and Lee}]{llava}
Haotian Liu, Chunyuan Li, Qingyang Wu, and Yong~Jae Lee. 2023.
\newblock Visual instruction tuning.
\newblock Available at: \url{https://arxiv.org/pdf/2304.08485}.

\bibitem[{Middlehurst et~al.(2021)Middlehurst, Large, Flynn, Lines, Bostrom, and Bagnall}]{middlehurst2021hive}
Matthew Middlehurst, James Large, Michael Flynn, Jason Lines, Aaron Bostrom, and Anthony Bagnall. 2021.
\newblock Hive-cote 2.0: a new meta ensemble for time series classification.
\newblock \emph{Journal of Machine Learning Research (JMLR)}.

\bibitem[{Middlehurst et~al.(2023)Middlehurst, Schäfer, and Bagnall}]{middlehurst2023bake}
Matthew Middlehurst, Patrick Schäfer, and Anthony Bagnall. 2023.
\newblock Bake off redux: a review and experimental evaluation of recent time series classification algorithms.
\newblock \emph{Journal of Time Series Analytics}.
\newblock In press.

\end{thebibliography}

\clearpage

\appendix
\section{Data Sample Examples}

\subsection{Baseline Representation}

The baseline representation shows the raw time-series signal in a comma-separated format. Below is an example of a data sample and the corresponding class prediction from the model.

\begin{figure}[H]
    \centering
    \includegraphics[width=1.0\linewidth]{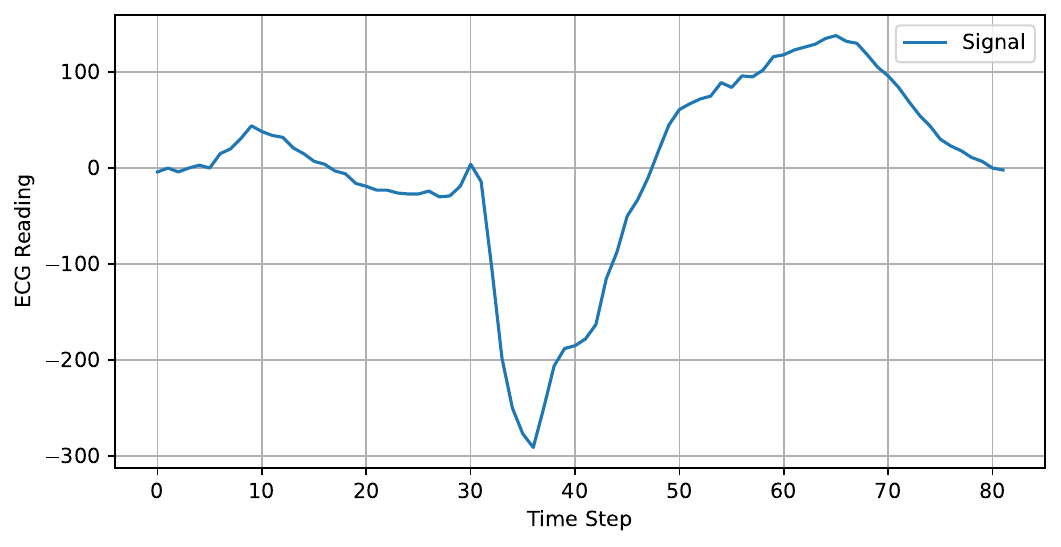}
    \caption{ECG Signal depicting amplitude across time-steps}
    \label{fig:ECG}
\end{figure}

\begin{quote}
\textbf{Human:} Which class is the following signal from? \\
\[
\begin{array}{cccccccc}
-4, & 0, & -4, & 0, & 3, & 0, \\
15, & 31, & 44, & 38, & 34, & 32, \\
21, & 15, & 4, & -3, & -6, & -16, \\
-19, & -23, & -23, & 69, & 55, & 44,\\ 
30, & 23, & 18, & 11, & 0, & -2
\end{array}
\]
\textbf{Model:} 1
\end{quote}

\subsection{WITH\_STATS Representation}

The WITH\_STATS representation shows the raw time-series signal in a comma-separated format augmented with summary statistics.

\begin{figure}[H]
    \centering
    \includegraphics[width=1.0\linewidth]{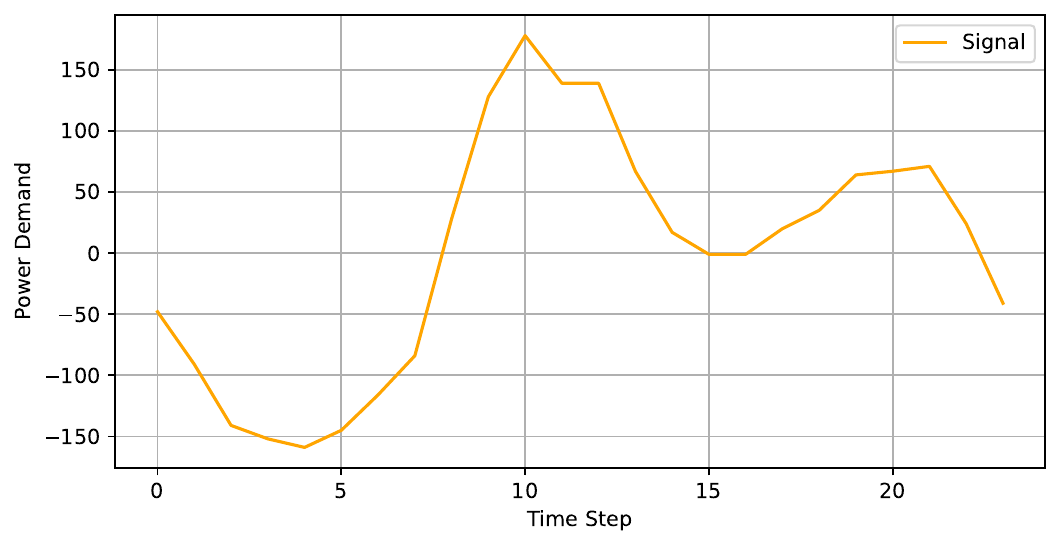}
    \caption{Power Demand across time-steps}
    \label{fig:Power Demand Signal depecting electrical power usage across time steps}
\end{figure}

\begin{quote}
\textbf{Human:} Which class is the following signal from? \\
\[
\begin{array}{cccccccc}
-125, & -141, & -156, & -158, & -161, \\
-143, & 50, & 92, & 107, & 94, \\
74, & 48, & 57, & 96, & 129, \\
105, & 63, & 34,
\end{array}
\]
\textbf{Summary Statistics:}
\begin{align*}
\text{Mean:} &\ 2.0833, \\
\text{Max:} &\ 1.2863, \\
\text{Min:} &\ -1.6148, \\
\text{Skewness:} &\ -5.7071, \\
\text{Entropy:} &\ 1.8891, \\
\text{Kurtosis:} &\ -1.2575
\end{align*}
\textbf{Model:} 1
\end{quote}

\subsection{Other Experiments: WITH\_RATIONALE}
In this approach, we further augment the time-series data with some rationales. These rationales are synthesized by prompting LLMs to justify why a certain classification makes sense. Our hypothesis here, was that a rationales-based approach ensures that the model is aligned with domain-specific knowledge. This can be particularly useful, when data is complex to interpret without clear justification . 

\begin{quote}
\textbf{Rationale Example:} The signal representation of the TwoLeadECG time series data exhibits distinct patterns characteristic of Class 1, which is typically associated with specific cardiac conditions or normal physiological responses. The initial fluctuations, including negative values and a gradual increase to a peak, suggest a typical depolarization and repolarization sequence observed in healthy heart rhythms. The presence of sharp transitions, particularly the significant drop to -102 and subsequent recovery, indicates a potential arrhythmic event or a response to stress, which aligns with the features of Class 1. Furthermore, the overall trend of the signal, with a return to positive values and stabilization towards the end, reinforces the notion of a transient event rather than a chronic condition, supporting the classification as Class 1. 

The combination of these features—initial negative deflections, a pronounced peak, and a return to baseline—demonstrates the signal's alignment with the expected characteristics of Class 1, confirming its classification.
\end{quote}

\subsection{Issues in Rationales Based Approach}
Our ability to generate Rationales did not generalize well across test-data . For a two-class classification problem, the similarity factor of the rationales turns out to be high, not really aiding in augmenting the predictive power of the model. This is obvious from the training plot of the rationales based approach. The training loss levels off to zero and stays there, characteristic of over-fitting to training data.  

\begin{figure}[H]
    \centering
    \includegraphics[width=1.0\linewidth]{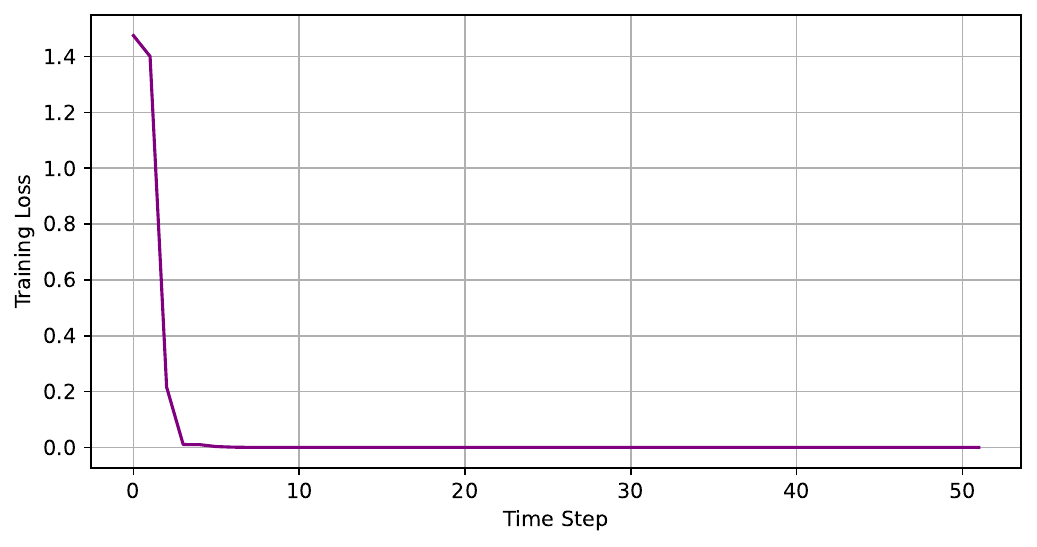}
    \caption{Training Loss across epochs for rationales-based approach}
    \label{fig:train_rationale}
\end{figure}

\subsection{Future Work}
The limitations of the rationale-based approach observed in this study highlight key areas for improvement. Future work will focus on:
\begin{itemize}
    \item Enhancing rationale diversity and generalization through dynamic selection techniques.
    \item Introducing regularization methods to mitigate over-fitting.
    \item Incorporating human feedback to refine and validate rationale generation.
    \item Developing task-specific metrics for evaluating rationale relevance and informativeness.
    \item Exploring integration with advanced architectures, such as attention mechanisms, to better utilize rationale information.
    \item Experimenting with alternative loss functions tailored to rationale quality and predictive power.
\end{itemize}

\end{document}